\documentclass[review]{elsarticle}

\usepackage{hyperref}

\usepackage[utf8]{inputenc}

\usepackage{amsmath}
\usepackage{amssymb}
\usepackage{xcolor}
\usepackage{graphicx}
\usepackage{geometry}
\usepackage{balance}
\usepackage{subcaption}
\usepackage{multirow}
\usepackage{styles}
\usepackage{tikz}
\usepackage{pgfplots}
\usepackage{filecontents}
\usepackage{placeins}
\usepackage{booktabs}
\usepackage{url}
\usepackage{wrapfig}
\usepackage{tikz}
\usetikzlibrary{arrows.meta}
\usepackage{adjustbox}
\usepackage{algpseudocode}
\usepackage{algorithm}

\journal{International Journal of Applied Earth Observation and Geoinformation}

\begin{document}

\begin{frontmatter}

\title{Predicting Vegetation Stratum Occupancy\\
from Airborne LiDAR Data with Deep Learning}

\author[1,2]{Kalinicheva Ekaterina}
\author[1]{Landrieu Loic}
\author[1]{Mallet Clément}
\author[1,3]{Chehata Nesrine}

\address[1]{LASTIG, Univ. Gustave Eiffel, IGN-ENSG, F-94160 Saint-Mandé, France}

\address[2]{INRAE, UMR 1202 BIOGECO, Université de Bordeaux, France}

\address[3]{EA G\&E Bordeaux INP, Université Bordeaux Montaigne, France}

\begin{abstract}
    We propose a new deep learning-based method for estimating the occupancy of vegetation strata from airborne 3D LiDAR point clouds. Our model predicts rasterized occupancy maps for three vegetation strata corresponding to lower, medium, and higher cover. Our weakly-supervised training scheme allows our network to only be supervised with vegetation occupancy values aggregated over cylindrical plots containing thousands of points. Such ground truth is easier to produce than pixel-wise or point-wise annotations. Our method outperforms handcrafted and deep learning baselines in terms of precision by up to $30$\%, while simultaneously providing visual and interpretable predictions. We provide an open-source implementation along with a dataset of 199 agricultural plots to train and evaluate weakly supervised occupancy regression algorithms.
\end{abstract}

\begin{keyword}
    vegetation analysis \sep
    stratum \sep
    airborne LiDAR\sep
    deep neural networks\sep
    weakly-supervised learning
\end{keyword}

\end{frontmatter}


\section{Introduction}
\label{sec:intro}
Estimating the structure of vegetation is a crucial first step for many environmental and ecological applications~\cite{dnubenmire1959canopy, morsdorf2010discrimination, bergen2009remote}.  
This is typically a time-consuming undertaking, often performed with \textit{in situ} visual approximate measurements \cite{willem2000ocular}.
Knowledge about the structure of vegetation is helpful for pasture land management \cite{velthof2014grassland}, and helps to better model the risk of forest fire \cite{mckenzie201115, maclean1996forest, sandberg2001characterizing}.

The progress in hardware precision and portability allows public and private actors to gather large quantities of geometric and radiometric data from airborne platforms \cite{chen2007airborne}. Such data sources are particularly well suited for vegetation analysis \cite{secord2007tree, STRIMBU201530,Lidar_detection_of_individual_tree}.
{Bolstered by the compelling performance \cite{guo2020deep} and increasing accessibility \cite{chaton2020torch} of deep learning for 3D point cloud analysis, we propose a deep learning approach to the problem of stratum occupancy prediction at the plot level.}
Our network can output two-dimensional stratum occupancy maps for three vegetation strata relevant to land pasture management: lower, medium, and higher vegetation.
{Each of these stratum occupancy maps is presented in a form of a regular grid---a rasterized map---of circular shape with a user-defined pixel size. Our algorithm is developed for $10$m-radius plot-level data with the purpose of using the final model for  large-scale mapping at the parcel level by dividing the area of interest into habitual plot-size samples.
}
A benefit of our approach is that it can be entirely supervised with values describing the average stratum occupancy over cylindrical plots, which can contain thousands of points. Such \emph{aggregated} values are much easier to produce than point-wise or pixel-wise annotations.

\paragraph{\bf Automated Vegetation Analysis} Vegetation analysis covers multiple tasks, depending on the level of analysis (tree-based, stand-based, plot-based, etc) and the area of interest (urban VS natural environments). During the two last decades, remote sensing has shown to be the most suitable solution for automatic information extraction \cite{HILDEBRANDT199055,PEKKARINEN2009171,LECHNER2020405,COOPS2021112477}, such as individual tree detection \cite{REITBERGER2009561, hyyppa2001segmentation, Lidar_detection_of_individual_tree,VEGA201498, f9120759, STRIMBU201530}, tree species classification \cite{DECHESNE2017129, Diedershagen04automaticsegmentation}, and structural and biophysical analysis of vegetation \cite{LEFSKY1999339,Generalizing_predictive_models_of_forest_inventory,Latifi2016EstimatingOA, 7762140}. 
Canopy analysis at the tree level, which entails biomass estimation, is typically conducted by combining the characteristics of individual trees. However, small trees are often missed by segmentation algorithms, leading to a less precise estimation of understory cover \cite{Williams20203DSO}. Therefore, several works propose to focus on modeling the understory layer explicitly \cite{venier2019modelling, campbell2018quantifying, wing2012prediction} or as one of several vegetation layers \cite{Latifi2016EstimatingOA}.
Several works also propose to characterize the stratification of vegetation \cite{morsdorf2010discrimination, ferraz20123}.
In the present paper, our objective is to automatically derive two-dimensional occupancy maps for different vegetation strata.


\paragraph{\bf Use of 3D LiDAR Sensors in Forestry}
The emergence of high-performing and compact LiDAR sensors has increased significantly the use of 3D data obtained from aerial platforms \cite{hyyppa2001segmentation, STRIMBU201530, Lidar_detection_of_individual_tree}. It has enabled operational forest mapping and inventory both at local and national scales \cite{Naesset07}. Indeed, contrary to optical images that only capture the upper vegetation layer, LiDAR is able to penetrate the tree canopy and provide precise geometric information about the vegetation structure for different strata.
Some works have combined LiDAR 3D point clouds with aerial images \cite{ke2010synergistic}, forest-centric GIS \cite{Diedershagen04automaticsegmentation}, or expert information on forest habitats \cite{Latifi2016EstimatingOA} in order to improve information extraction. Focusing on an operational and reproducible scenario, our approach operates on a common acquisition setting in which a LiDAR acquisition is combined with a simultaneous multi-spectral very high resolution optical acquisition. This leads to the generation of a 3D point cloud attributed with both geometric and radiometric information.

\paragraph{\bf Traditional Approaches to Vegetation Structure Analysis}
In order to exploit the rich structural information of aerial LiDAR scans, researchers have developed two main approaches. 
The \emph{Area-based} approach consists in 
deriving handcrafted descriptors from 3D acquisitions and regressing 
vegetation features for a subset of acquisition \cite{Latifi2016EstimatingOA, Generalizing_predictive_models_of_forest_inventory}.
Such prediction models have been adopted for operational purposes since they require lower point densities and are computationally efficient. However, they require a sufficient amount of ground-based measurements for establishing reliable models \cite{rs2061481}.
The \emph{Tree-based} first delineates individual trees and in turn aggregates morphological indicators across the area of interest. Such methods typically start with a non-parametric detection method 
\cite{HAMRAZ2016532} and then use clustering algorithms like watershed \cite{Chen2006IsolatingIT}, region growing \cite{hyyppa2001segmentation}, or graph-based methods \cite{STRIMBU201530, REITBERGER2009561}. {A limitation of these segmentation task it tends to overlook small subdominant trees and focus on larger trees, limiting subsequent stratum analysis.} 
As our method directly operates on a plot and does not require any pre-segmentation step, it qualifies as \textit{Area-based}. However, we produce rasterized occupancy maps whose pixel size is typically sub-metric.

\paragraph{\bf Learning-Based Vegetation Structure Analysis}
In order to achieve high generality without requiring expert knowledge on the vegetation structure of the considered area, several works have explored the benefit of learning-based approaches for automated forest analysis, see the review of Liu \etal \cite{liu2018application}. 
More recently, the first deep learning methods operating on 3D forestry data have been proposed. Lang \etal investigate the possibility of achieving global-scale mapping with satellite-borne LiDAR and Bayesian deep learning \cite{lang2021global}.
Several approaches based on 2D convolutional networks have been proposed for the classification of individually segmented trees  \cite{zou2017tree, seidel2021predicting, Deep_learning_for_conifer_deciduous, chen2021individual}.
{However, using networks designed for 2D to analyse 3D data not only incurs costly pre-processing steps, but generally leads to lower performance than using architectures dedicated to 3D data\cite{guo2020deep}.}
Furthermore, these methods require  databases with precisely segmented trees, which makes them less applicable to the operational setting of vegetation structure prediction. This is particularly problematic for natural forests for which tree' canopies often intersect, making segmentation a difficult and sometimes dubious process.

\paragraph{\bf 3D Deep Learning} The main difficulty in analysing 3D point clouds with deep learning is their irregular structure and varying sampling density.
This has been alleviated by relying on images \cite{boulch2018snapnet, MVCNN, view_gcn}, 
3D regular grids \cite{octnet, voxnet, choy20194d, graham20183d}, graphs \cite{simonovsky2017dynamic, landrieu2018large}, or continuous-space convolutions \cite{boulch2020convpoint, kpconv}.
A simpler class of algorithms considers point clouds as unordered sets of points \cite{pointnet, pointnet2, zaheer2017deep}, and does not need any of the pre-processing steps required by the aforementioned methods.
Since we consider the strata occupancy prediction problem for plots individually, and with the objective of scalability and computational efficiency, we choose the straightforward PointNet network \cite{pointnet}. 


\paragraph{\bf Weakly Supervised Learning}
Training deep networks typically requires a large training database. However, manually producing dense annotations of LiDAR 3D point clouds of vegetation is a laborious task, often made even more complicated by visual ambiguities \cite{milberg2008observer}. Furthermore, since most monitoring tasks are done at the plot level, such a degree of detail is unnecessary in practice. Similarly to what Long \etal  developed for images \cite{tong2021point}, our approach only requires sparse annotations to be trained. Our network can be entirely supervised from a single aggregated occupancy per cylindrical plot and per stratum.
Such values can be obtained by an operator estimating visually the stratum occupancy of their immediate surrounding area. While this requires an in-situ intervention, the annotation task is less tedious than annotating individual points.
As commonly encountered in weakly supervised schemes \cite{ratner2019weak,liu2019weakly}, our method requires regularization terms for more realistic outputs.

The key contributions of this paper are as follows:
\begin{itemize}
    \item We show that a simple deep network can produce two-dimensional stratum maps using only plot-aggregated weak annotation.
    \item We propose regularization terms improving the realism and generality of the predicted occupancy maps without requiring expert knowledge.
    \item We introduce an open-access dataset of multi-spectral 3D point clouds corresponding to plots of agricultural parcels along with aggregated stratum occupancy annotations.
\end{itemize}

\section{Materials and methods}

\begin{figure}[ht]
    \centering 
\begin{tabular}{ccc}
\begin{subfigure}{0.17\textwidth}
  \includegraphics[width=\linewidth, height=.17\textheight]{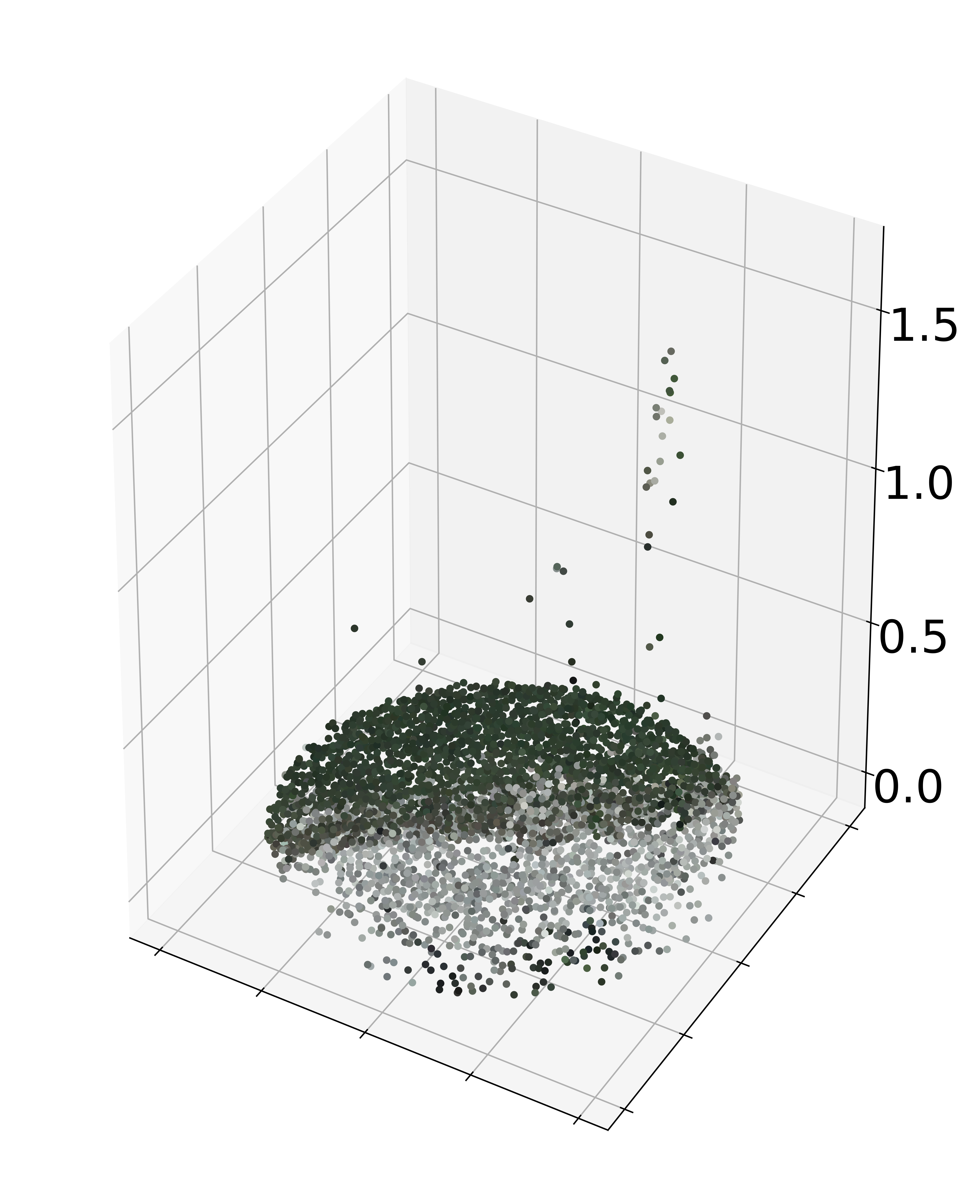}
\end{subfigure}&
\begin{subfigure}{0.25\textwidth}
  \includegraphics[width=\linewidth, height=.17\textheight]{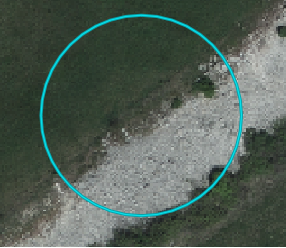}
\end{subfigure}&
\begin{subfigure}{0.51\textwidth}
  \includegraphics[width=\linewidth, height=.17\textheight]{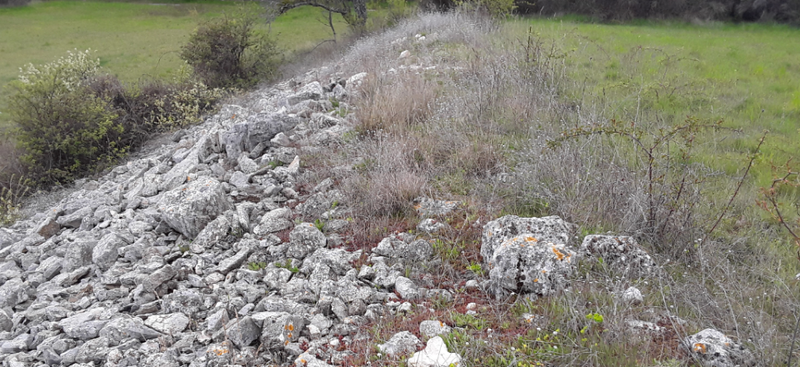}
\end{subfigure}\\
\begin{subfigure}{0.17\textwidth}
  \includegraphics[width=\linewidth, height=.17\textheight]{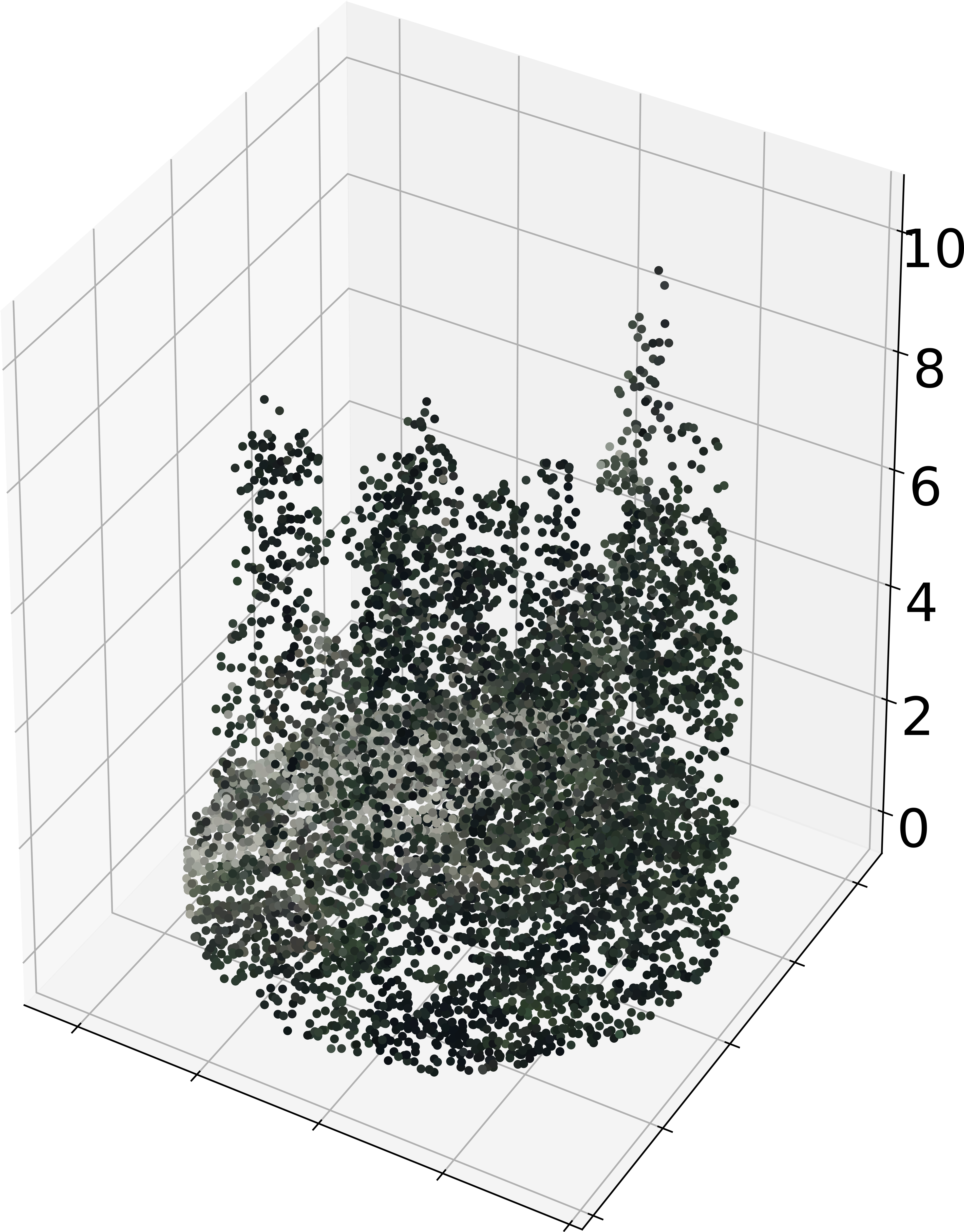}
  \caption{LiDAR scan.}
  \label{fig:lidar}
\end{subfigure}&
\begin{subfigure}{0.25\textwidth}
  \includegraphics[width=\linewidth, height=.17\textheight]{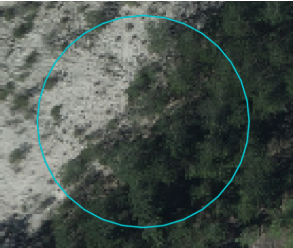}
  \caption{Aerial image.}
  \label{fig:camera}
\end{subfigure}&
\begin{subfigure}{0.51\textwidth}
  \includegraphics[width=\linewidth, height=.17\textheight]{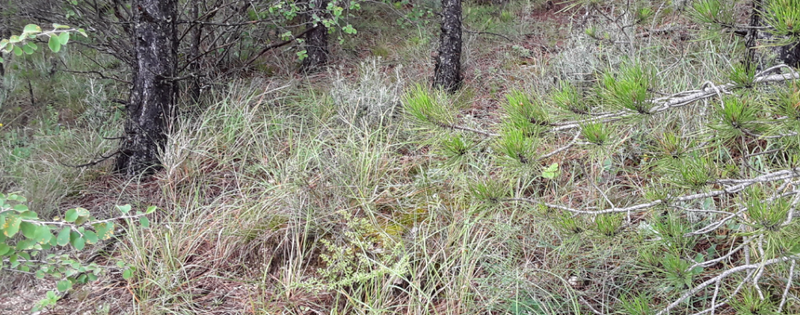}
  \caption{In situ view.}
  \label{fig:situ}
\end{subfigure}
\end{tabular}   
\caption{{\bf Examples of plot-based acquisitions.} 
The two point clouds in \Subref{fig:lidar} correspond to two distinct plots. They are colored using the aerial images in \Subref{fig:camera} (the plots are represented by the circles). A human annotator visually assesses their surroundings \Subref{fig:situ} and estimates the occupancy ratio of the lower, medium, and higher vegetation stratum. In both represented plots, the occupancy ratio of the lower stratum is $50\%$, while the occupancy higher strata differs.}
\label{fig:study_plots_photos}
\end{figure}

\subsection{Dataset Description}
\label{sec:data}

In this subsection, we present the new proposed dataset and give further precision on the task of automated stratum occupancy prediction.

\paragraph{\bf Dataset Composition} We gathered a total of $199$ aerial LiDAR scans of cylindrical plots with a $10$m radius with an average $10$-pulse per square meter density. The plots have been selected by forestry experts, and correspond to typical pasture land parcels in South-Eastern France (see \figref{fig:study_plots_photos}). A RGB-IR camera sensor captures Red-Green-Blue- Near InfraRed radiometric information simultaneously with the LiDAR acquisition.

Each plot comprises between $3\,000$ to $17\,000$ 3D points, and each point is attributed with a total of $9$ features: (i) absolute 3D coordinates in Lambert-$93$ system
, (ii) RGB and Near-InfraRed reflectance values acquired by the aerial camera, (iii) uncalibrated laser intensity and return number as provided by the LiDAR device. Since there is no theoretical framework proposing intensity calibration over forest plots, we decided to keep the raw values. 

In order to align the 3D point clouds and the camera acquisition, we simply project the pixels' colors to the corresponding 3D points without taking occlusion into account.
We could also have added the total number of returns for each LiDAR ray, but observed that this information is in practice redundant with the return number.

\begin{figure}[h]
\captionsetup[subfigure]{justification=centering}
    \centering
    \begin{tabular}{cc}
    \adjustbox{valign=m}{\begin{tabular}{c@{}c@{}}

    \begin{subfigure}{0.5\textwidth}
        \centering
        \includegraphics[scale=0.6]{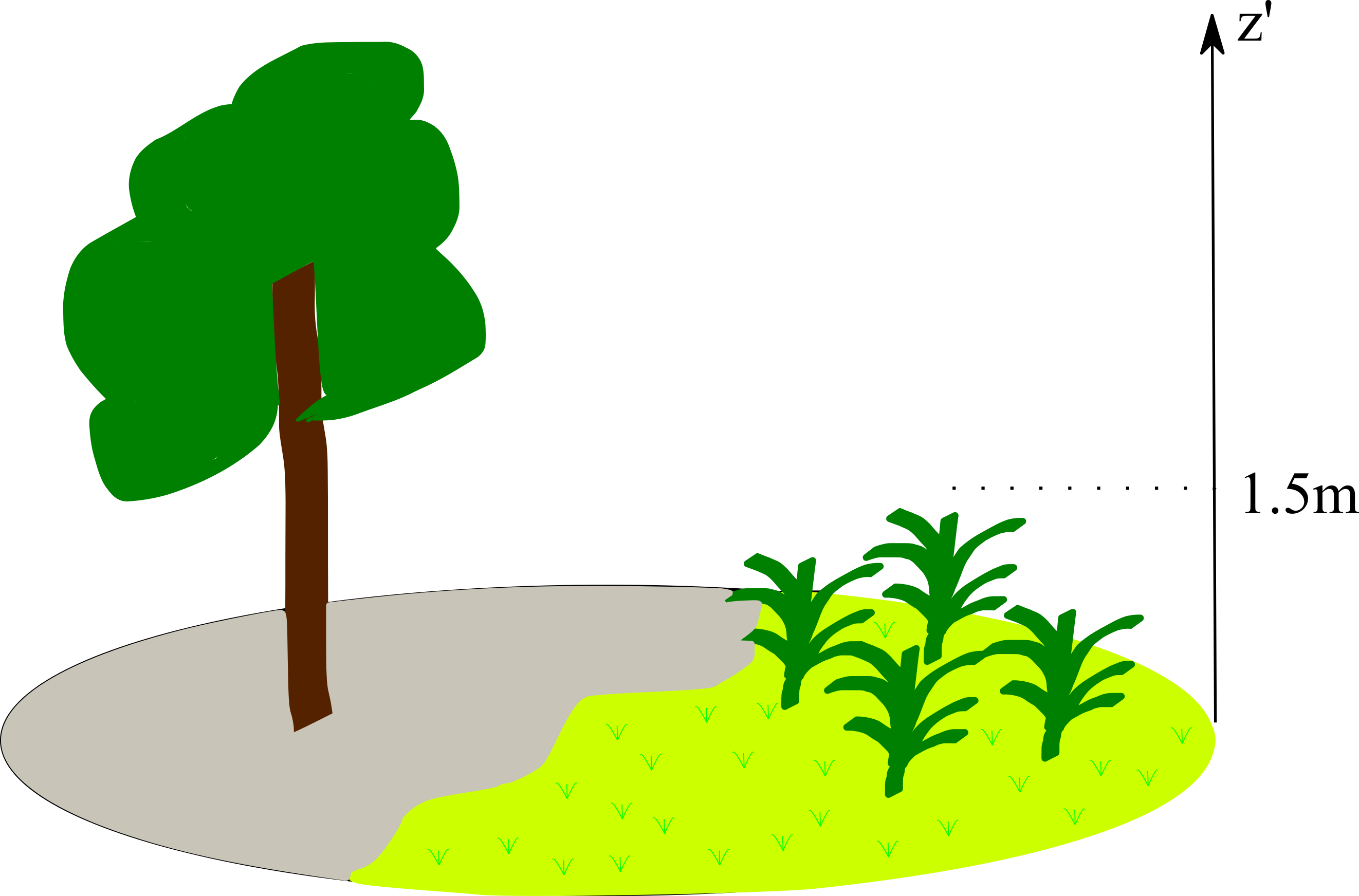}
        \caption{Study Plot.}
        \label{fig:plot_coverage:a}
    \end{subfigure}
    \\
    \begin{subfigure}{0.5\textwidth}
    \centering
        \includegraphics[scale=0.6]{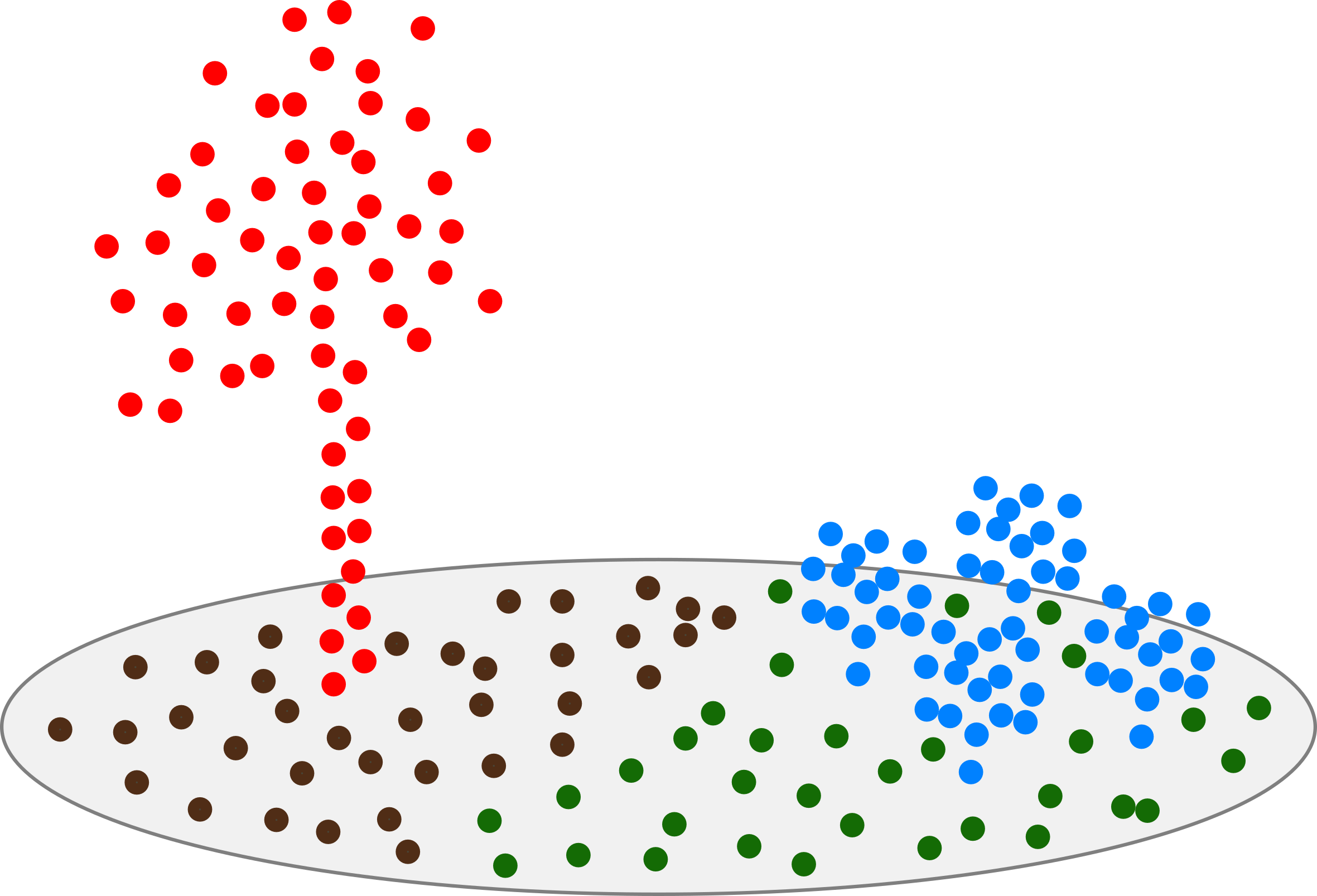}
        \caption{Pointwise labelling:\\ \textcolor{ brown!50!black}{\bf soil}, \textcolor{green!50!black}{\bf grass}, \textcolor{blue!70!cyan}{\bf medium vegetation}, \textcolor{red}{\bf high vegetation}.}
        \label{fig:plot_coverage:b}
    \end{subfigure}
    \end{tabular}}
         &
    \adjustbox{valign=m}{\begin{tabular}{c@{}c@{}c@{}}
    \begin{subfigure}{0.45\textwidth}
    \centering
        \includegraphics[scale=0.6]{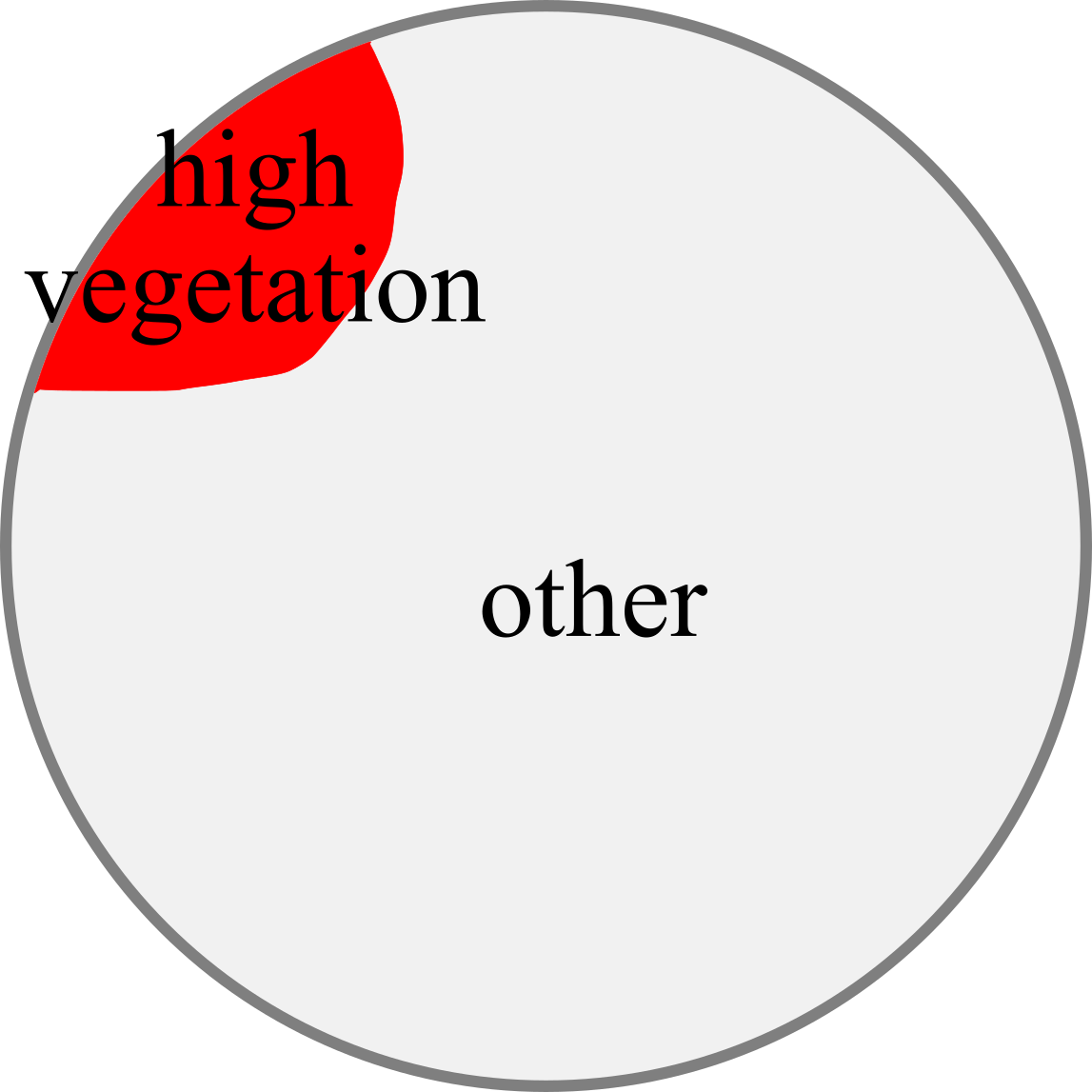}
        \caption{Higher stratum: $o_\HL $= 10\%.}
        \label{fig:plot_coverage:c}
    \end{subfigure}
         \\
    \begin{subfigure}{0.45\textwidth}
    \centering
        \includegraphics[scale=0.6]{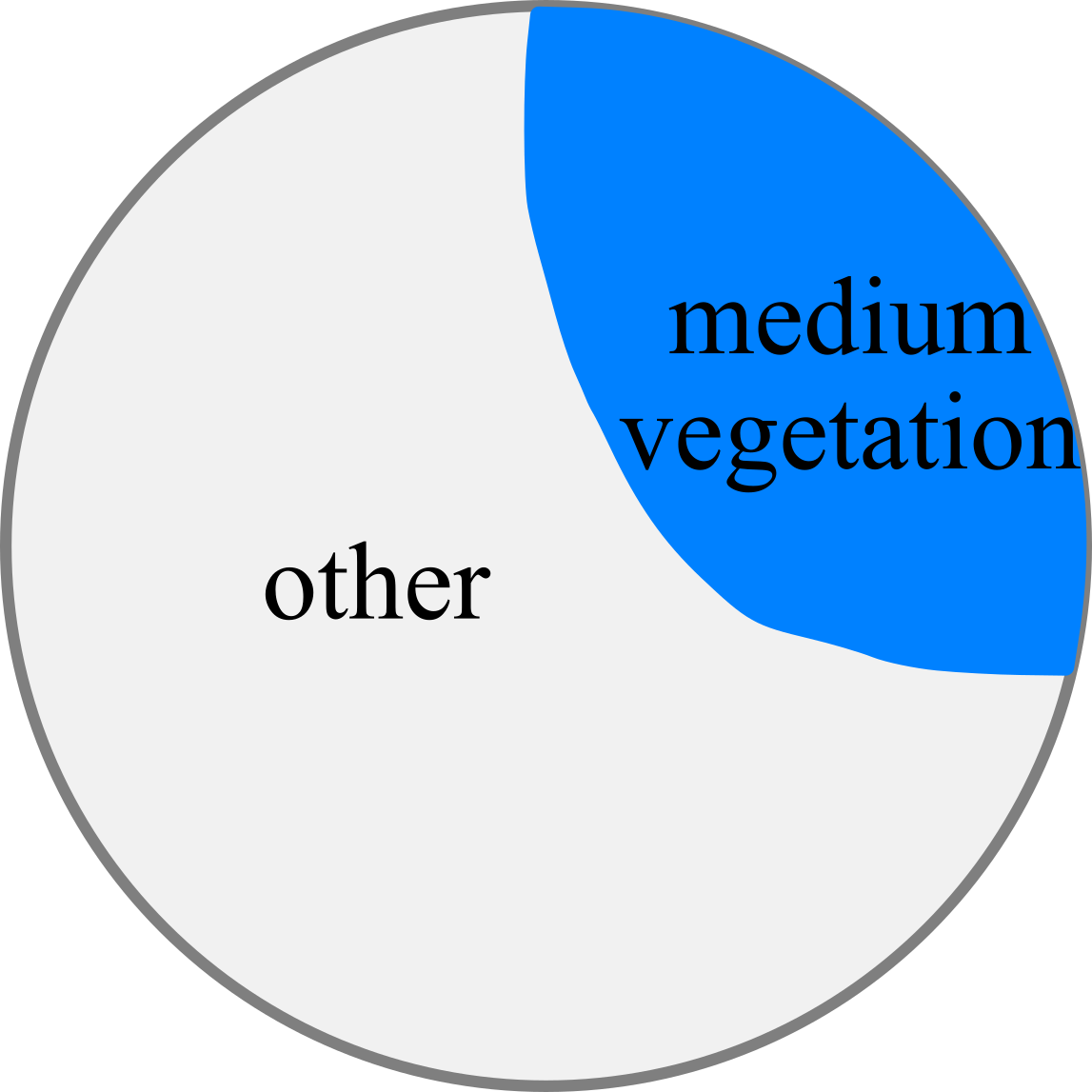}
        \caption{Medium stratum: $o_\ML $= 25\%.}
        \label{fig:plot_coverage:d}
    \end{subfigure}
         \\ 
    \begin{subfigure}{0.45\textwidth}
    \centering
        \includegraphics[scale=0.6]{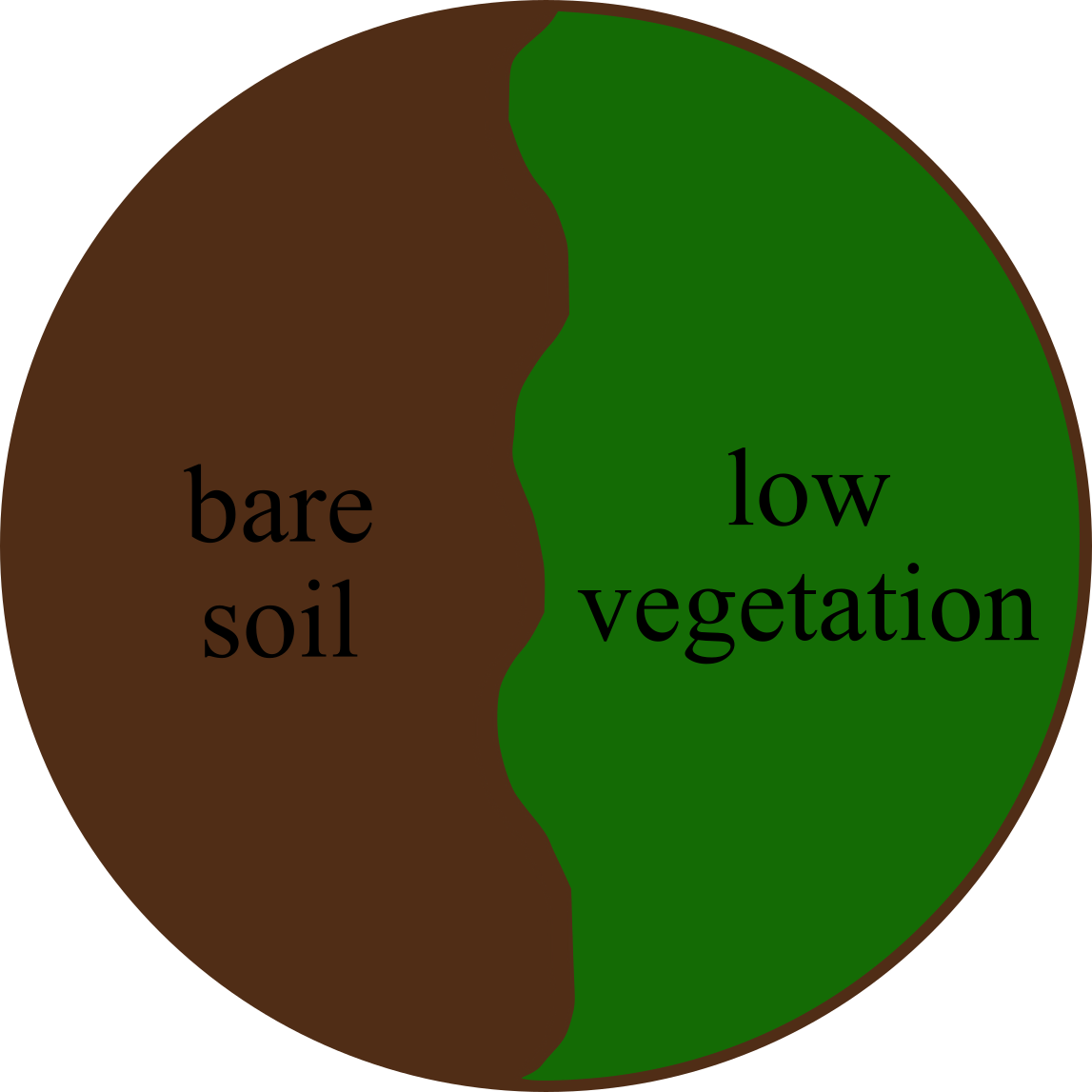}
        \caption{Lower stratum: $o_\LL $ = 50\%.}
        \label{fig:plot_coverage:e}
    \end{subfigure}
    \end{tabular}}
    \end{tabular}
    \caption{\textbf{Occupancy-annotated plot.} In \Subref{fig:plot_coverage:a}, we represent a synthetic scene whose point labels are represented in \Subref{fig:plot_coverage:b}. We illustrate the occupancy of the higher, medium, and lower stratum in \Subref{fig:plot_coverage:c}, \Subref{fig:plot_coverage:d} and \Subref{fig:plot_coverage:e}, respectively. The lower stratum is covered in equal proportions by low vegetation and bare soil, while the medium stratum is occupied at $25\%$ by bushes and the higher stratum is occupied at $10\%$ by the tree crown. Note that the lower part of the tree trunk represented in red in \Subref{fig:plot_coverage:b} is not counted as medium vegetation despite being under $1.5m$.}
    \label{fig:plot_coverage}
\end{figure}

\paragraph{\bf Normalisation}
The $x$ and $y$ values of the points in each plot are normalized within the unit square $[-1,1]^2$. 
The $z$-value---or height---of each point is normalized locally by subtracting the z-value of the lowest point in a $0.5$m cylindrical neighborhood. This simple approach allows to compensate for irregular terrain and avoid the propagation of potential errors in the Digital Terrain Model \cite{MALLET2011S71}.

\paragraph{\bf Annotation} Each plot has been annotated by a human expert \emph{in situ}, as illustrated in \figref{fig:study_plots_photos}. The annotation describes the occupancy of different strata of the cylindrical plot as assessed visually by the human annotator. More precisely, we are provided with the lower stratum occupancy ratio~$\hat{o}_\LL$, the medium vegetation stratum occupancy ($\hat{o}_\ML$), and the higher strata occupancy ($\hat{o}_\HL$), see \figref{fig:plot_coverage}. The occupancy value $\hat{o}_\LL$ characterizes the proportion of the ground surface occupied by grass or low vegetation, as opposed to stone, soil, or sand. $\hat{o}_\ML$ characterizes the proportion of the surface of the plot occupied by the footprint of medium vegetation, \ie with a 
height between $0.5$ and $1.5$m.
This type of vegetation, typically bush-like, is the most accessible by pasture animals and represents an important indicator for land-use monitoring agencies. Note that the trunks of trees exceeding $1.5$m do not not contribute to this coverage.
Finally, the canopy occupancy is defined as the ratio of the plot surface occupied by the footprint of the canopy of trees over $1.5$m.

Our dataset \cite{ekaterina_kalinicheva_2021_5555758} is publicly available at \url{https://doi.org/10.5281/zenodo.5555758}.


\subsection{Weakly-Supervized Stratum Prediction}
\label{sec:method}
\begin{figure}[t]
    \centering
    \includegraphics[scale=0.82]{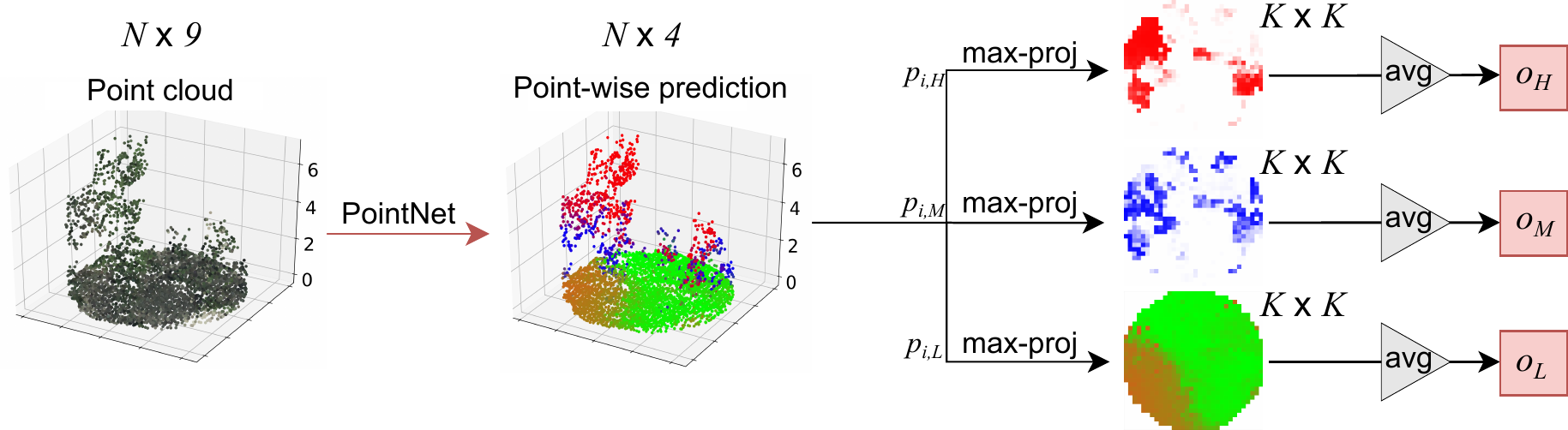}
    \caption{\textbf{Pipeline.} Our network performs the semantic segmentation of a 3D point cloud within four different classes. The resulting probabilities are projected onto rasters corresponding to different strata. Finally, the occupancy maps are aggregated into the stratum vegetation ratio.}
    \label{fig:pointnet_bay}
\end{figure}

We consider a point cloud $X \in \bfR^{N \times 9}$ with $N$ points. Each point is characterized by the $9$ radiometric and geometric features described in \secref{sec:data}. Our objective is to predict rasterized vegetation occupancy maps of the three vegetation strata (Figure~\ref{fig:pointnet_bay}): lower ($o_\LL$), medium ($o_\ML$), and higher ($o_\HL$). We first predict a semantic class for each point (\secref{sec:pointwise}), then aggregate these predictions into explicit rasterized stratum occupancy maps (\secref{sec:stratum_modeling}). 
{To improve our weakly-supervised model, we use some regularization terms. We introduce in \secref{sec:bayesian} an unsupervised elevation model encouraging more coherent classification.
In \secref{sec:entropy}, we present our entropy--based prior designed to produce crisper maps. All can be incorporated into a global loss function presented in \secref{sec:loss}}

\subsubsection{Point-wise Class Prediction}
\label{sec:pointwise}
We classify each point among $C=4$ classes: lower vegetation, bare soil, medium vegetation, and higher vegetation. 
 For this task, we use the straightforward PointNet semantic segmentation network \cite{pointnet}, see \ref{sec:pointnet} of the appendix.

\begin{figure}[t]
    \centering
    \includegraphics[scale=1]{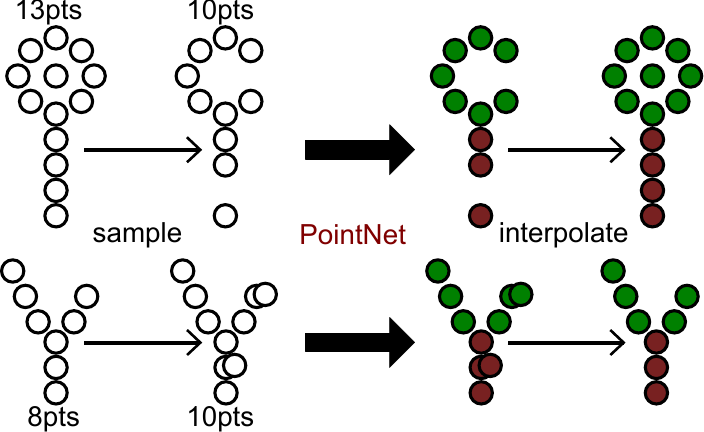}
    \caption{\textbf{Point Sampling.} Each point cloud is sampled to the same size $M$. When $N<M$, we duplicate randomly chosen points to get the desired point count. Then, the sampled points are classified with a PointNet network. Finally, the predictions are interpolated to the full size of the initial point.}
    \label{fig:downsampling}
\end{figure}

In order to handle the varying density of point clouds and to facilitate batch-training, we first sample each plot's point cloud $X$ into the same number $M$ of points. The sampled points are classified by a PointNet network, and in turn interpolated to the initial full size point cloud with nearest neighbor interpolation, see \figref{fig:downsampling}.
%
We denote the predicted probabilities for a point $n\in[1, N]$ as follows: ($Y_{n,\text{BS}}$) for bare soil, ($Y_{n,L}$) for low, ($Y_{n,M}$) for medium and ($Y_{n,H}$) for high vegetation, respectively.
\subsubsection{Stratum Modeling}
\label{sec:stratum_modeling}

In order to obtain stratum occupancy maps and predictions, we project the point cloud onto the stratum rasters and aggregate the function into a ratio prediction which we can supervise end-to-end.

\begin{figure}
     \centering
    \includegraphics[width=.4\textwidth]{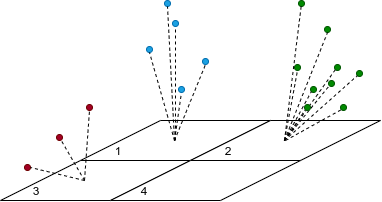}
    \caption{\textbf{Projection onto Stratum Rasters.} Example of 3D data projection to the 2D space. A different number of points can be projected to one pixel, or possibly none  (pixel 4) depending on point density of the area.}
    \label{fig:projection}
\end{figure}

\paragraph{\bf Point Projection} We use point-wise prediction to estimate the occupancy of pixels of each stratum. We consider a raster of $K \times K$ pixels aligned with the projection of the cylindrical plot on the horizontal axes. We associate to each pixel $(i,j)$ of the raster with the set of 3D points $\text{proj}(i,j) \subset [1, \cdots, N]$ whose vertical projection \emph{falls} in the pixel's extent:
\begin{align}
    \text{proj}(i,j) = 
    \left\{
    n \in [1,N] \middle| 
    \left\lfloor\frac{x_n}{K}\right\rfloor=i,
    \left\lfloor\frac{y_n}{K}\right\rfloor=j
    \right\}~,
\end{align}
with $x_n$ and $y_n$ the $x$ and $y$ coordinate of a point $n$. Note that the cylindrical shape of the plot implies that only pixels within a disk inscribed in the raster will be associated with any point, see \figref{fig:projection}.

\paragraph{\bf Stratum Aggregation}
We compute an occupancy $O^{(\text{stratum})}_{i,j}$ for each pixel  $i,j\in [1,K]^2$ and each stratum by taking the highest predicted probability for all points associated with this pixel. Finally, we aggregate the pixel projection stratum-wise to obtain a single prediction for each plot and stratum. For each $\text{stratum}$ in $\{L,M,H\}$ we have:
\begin{align}
\label{eq:pixel_value}
    O^{(\text{stratum})}_{i,j} &= \max_{n \in \text{proj}(i,j)} Y_{n, \text{stratum}}
    \\
    o_\text{stratum} & = \frac1D \sum_{i,j=1\cdots k}  O^{(\text{stratum})}_{i,j}  
\end{align}
with $D$ the number of pixels within the disk obtained when projecting the cylindrical scan ($D\sim~\pi/4 K^2$). Note that this projection raster allows us to visualize pixel-precise predicted occupancy maps for each stratum.

\paragraph{\bf Occupancy Supervision}
We supervise the predicted stratum occupancies with the ground truth annotated occupancies $\hat{o}_\LL$,$\hat{o}_\ML$,$\hat{o}_\HL$ using the distance function $\phi$:
\begin{align}
\label{eq:loss_l1}
\mathcal{L}_{data} = \phi({o}_L - \hat{o}_L) + \phi({o}_M - \hat{o}_M)+ \phi({o}_H - \hat{o}_H)~.
\end{align}
In practice, we use $\phi(x) = \sqrt{x^2+\scriptsize{10^{-4}}}$ as a differentiable surrogate of the $\ell_1$ norm.
\subsubsection{Elevation Modeling}
\begin{figure}
    \centering
    \begin{tikzpicture}
    \begin{axis}
    [xmin=0,xmax=9,
	 ymin=0,ymax=0.3,
	 xlabel=elevation (m), ylabel=probability density]
    \addplot [ybar, bar width=2pt, color=blue,fill=blue] table[col sep=comma,header=false, x index = 0,y index=1] {./images/ECM.csv};
    \addlegendentry{\small{Empirical distribution}}
    \addplot [color=green!80!black, ultra thick] table[col sep=comma,header=false, x index = 0,y index=2] {./images/ECM.csv};
    \addlegendentry{\small{Ground component: $\Gamma(0.18,0.49)$}}
    \addplot [color=red, ultra thick] table[col sep=comma,header=false, x index = 0,y index=3] {./images/ECM.csv};
    \addlegendentry{\small{Non-ground component: $\Gamma(2.19,2.50)$}}
     \end{axis}
   \end{tikzpicture}
    \caption{\textbf{Elevation Modeling.} Empirical elevation distribution (blue) and two components of a fitted mixture of Gamma distributions with weight $0.55$ and $0.45$. The green component models the elevation of ground and low vegetation, while the red component models the medium and high vegetation.}
    \label{fig:bayesian}
\end{figure}
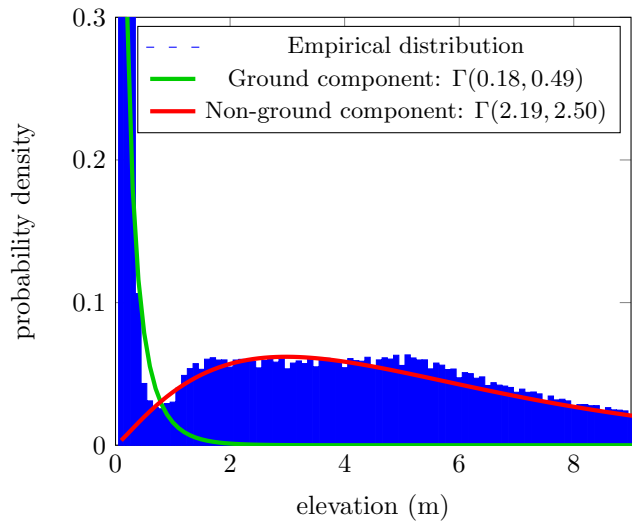

\label{sec:bayesian}
The model described above does not explicitly model the distribution of elevations within each stratum. In theory, points several meters above the ground can contribute to the lower vegetation stratum as long as the stratum-wise aggregated values are in agreement with the ground truth. We propose to explicitly model the elevation of points within each stratum in an unsupervised way with the goal of making the occupancy maps more realistic, and to increase the generalizability of the models.

By plotting the elevation histograms of all points as seen in \figref{fig:bayesian}, we observe that this empirical distribution follows a mixture model of two Gamma distributions. Moreover, we can easily interpret its components: the low elevation density peak corresponds to bare soil and low vegetation, while the long-tailed high elevation distribution corresponds to medium and high vegetation. To simplify the problem, we group the three strata into two groups: ground $\G$ and non-ground $\NONG$ (medium and high vegetation).

We can estimate the parameters $\{\scale_i,\shape_i\}_{i \in \G,\NONG} \in \mathbf{R}^2$ of both Gamma distributions as well as the mixture parameter $\{\rho_\G, \rho_\NONG\} \in [0,1]^2$ with the expectation–conditional–maximization (ECM) algorithm \cite{young2019finite} detailed in \algoref{alg:ECM}. This procedure is entirely unsupervised but requires a meaningful initialization of mixture parameters, which can be achieved by trial-and-error guided by the resulting likelihood value. The ECM algorithm and its inner Newton-Rachson optimization \cite[Chapter 3.2]{cheney2012numerical} converges in only a few iterations, and this step is only required to be performed once per dataset. In \figref{fig:bayesian}, we represent the mixture of Gamma distributions obtained with the ECM algorithm. We denote by $\Gamma_\G$ and $\Gamma_\NONG$ the distributions whose parameters (scale and shape) are learnt with the ECM algorithm. 
\begin{algorithm}
\caption{ECM Algorithm for Gamma Mixture estimation.\\
$\psi$ denotes the digamma function
     \cite[Chap. 6.3]{abramowitz1948handbook}}
     \label{alg:ECM}
\begin{algorithmic}
\Require Input: elevations $z \in \bfR_+^N$
\Require   $\scale_i$, $\shape_i$, $\rho_i$ $\gets$ manual initialization for $i \in\{\G, \NONG\}$
\While{not converged} 
    \For{$n \in [1,N], i \in\{\G, \NONG\}$}\Comment{E-step}
    \State $\gamma^i_n = \rho_i \Gamma(z_n; \scale_i, \shape_i)$ 
    \State $e^i_n = \gamma^i_n / (\sum_{j \in \G, \NONG} \gamma^j_n)$ \Comment{Expectation that point $n$ is in group $i$}
    \EndFor
    \For{$i \in\{\G, \NONG\}$}\Comment{Conditional M-steps}
        \State $\rho_i \gets \frac1N \sum_{n=1}^N e^i_n$ \Comment{mixture parameter} 
        \State $\alpha_i \gets \text{root of} \sum_{n=1}^N \gamma_i^n [\log(z_n) + \log(\shape_i) - \psi(\scale_i)]$  \Comment{With Netwton-Rachson}
        \State $\beta_i \gets {N \alpha_i \rho_i}/{\sum_{n=1}^N z_n e^i_n}$
     \EndFor
\EndWhile 
\end{algorithmic}
\end{algorithm}
As described in the Appendix, the Bayesian theorem allows us to define loss $\ell_\text{elevation}$ as the negative log-likelihood of the observed elevations $Z \in \mathbf{R}^N_+$ conditionally to the observations $X \in \mathbf{R}^{N \times 9}$:
\begin{align}
\mathcal{L}_\text{elevation}
&=-\sum_{n=1}^N \log\left(
     (Y_\LL + Y_\BS) \Gamma_\G(z_n)
      + (Y_\ML + Y_\HL) \Gamma_\NONG(z_n) \right)~.
\end{align}
This function encourages the network to classify points with low elevation as ground or low vegetation, and points with high elevation as medium or high vegetation. While this could arguably be done by setting a manual threshold or hand-picked parameters, this method is adaptive to each new dataset and converges quickly.
\subsubsection{Occupancy Prior Modeling}
\label{sec:entropy}
Pixel values of the vegetation occupancy maps take continuous values from $0$: no vegetation, to $1$: completely covered by vegetation.
This value can also be influenced by the confidence of the point classifier: an ambiguous pixel with low confidence may be predicted at $0.5$ occupancy. Once aggregated plot-wise, such indecisive predictions may average to the correct prediction and lead to a low loss. However, the resulting stratum occupancy maps may become fuzzy and hard-to-interpret. We would like to reserve values such as $0.5$ for the rare case of pixels which are partially covered by a given vegetation structure.
In order to discourage the network to express its uncertainty through intermediate prediction, we propose to regularize our loss with the average entropy of the pixel prediction of all maps:
\begin{equation}
\mathcal{L}_\text{entropy} = -\frac1{D}
\sum_{\substack{\text{stratum} \\ \in \{L,M,H\}}}
\sum_{{(i,j)  \in [1,K]^2}}
H\left(O^{(\text{stratum)}}_{i,j}\right)~,
\end{equation}
with $H(\cdot)$ the function returning the entropy of an input distribution, and $D$ the number of pixels in the cylinder projection onto the raster.
\subsubsection{Global Loss}
\label{sec:loss}
We add the elevation loss $\mathcal{L}_\text{elevation}$ and entropy loss $\mathcal{L}_\text{entropy}$ as regularizers of the data loss $\mathcal{L}_\text{data}$. The resulting loss for model optimization is computed batch-wise and averaged over the batch's plots:
\begin{align}
    \mathcal{L} = \mathcal{L}_\text{data} + \lambda \mathcal{L}_\text{elevation} + \mu \mathcal{L}_\text{entropy}~,
\end{align}
with $\lambda=1$ and $\mu=0.2$ the respective regularization strengths of $\mathcal{L}_\text{elevation}$ and $\mathcal{L}_\text{entropy}$.

\subsection{Implementations Details}
Our entire pipeline is implemented in PyTorch $1.7$ and CUDA $10.2$. Our network is trained with a batch size of $20$ plots for $100$ epochs, and we use the ADAM  optimizer\cite{kingma2017adam} with a learning rate of $0.001$ divided by $10$ after 50 epochs, and all other default parameters.
We add a dropout layer \cite{JMLR:v15:srivastava14a} with probability  $0.4$ before the last layer to increase the model's robustness.
Our network can be trained in under 24 minutes on an NVIDIA GeForce RTX 3060 GPU and a Xeon W-2123 CPU with $64$GB of RAM. Our python impementation of the ECM algorithm converges in under $5$ seconds on a standard workstation for over $500\,000$ points.

During both training and inference, we sample a fixed number of $M=4096$ points for each plot, and duplicate points for plots with fewer points. This allows us to use efficient batch-parallel computing, as well as adding sampling stochasticity to decrease overfitting. The size of the stratum raster is set to $K=32$ pixels.

The layer sizes of three MultiLayer Perceptron (MLP) blocks of our PointNet model are respectively: $[32,32]$, $[64,128]$ and $[64, 32, 4]$, see \ref{sec:pointnet}.

\subsection{Experimental Setting}
We perform $5$-fold cross-validation on the dataset presented in \secref{sec:data}, which is composed of $T=199$ cylindrical plots. We report the mean absolute occupancy error $e_\LL, e_\ML, e_\HL$ between the predicted and true occupancy for each stratum, as well as the inter-stratum macro-average $e$:
\begin{align}
    e_K &= \frac1T \sum_{n=1}^T 
    \left\vert
    \hat{o}^K_n - {o}^K_n
    \right\vert\;\;\text{for}\;\; K \in \{\LL, \ML, \HL\} \\
    e &= \frac13 \left(e_\LL +  e_\ML + e_\HL\right)~.
\end{align}

Our code is available at \url{https://github.com/ekalinicheva/plot_vegetation_coverage}.




\section{Results and discussions}
\label{sec:experiments}
In this section, we present an experimental evaluation of the performance of our approach, compared to two baselines relying on handcrafted features and a simple deep network.

\subsection{Competing Approaches}
We propose two baseline approaches to assess the performance of our method: a classic tree-based algorithm operating on handcrafted features and a simple deep learning-based regressor.
\paragraph{\bf Handcrafted Algorithm}
This classic method is composed of several steps:
\begin{itemize}
    \item \textbf{Lower Stratum Occupancy.} 
    We first consider all points with a normalized elevation under $0.5$m. We then select the points of all plots with $0\%$ lower stratum vegetation occupancy and average the $6$ non-geometric normalized point features to form a prototypical \emph{bare soil} point. Likewise, we form a prototypical \emph{low vegetation} point. We then classify all points according to their Euclidean distance with respect to the bare soil and low vegetation prototypes.

    Finally, all points with an elevation below $0.5$m are projected onto the raster corresponding to the lower stratum, and the pixels are classified as bare soil or low vegetation by a majority vote. This allows us to compute a predicted occupancy for the lower stratum.
    \item \textbf{Medium and Higher Stratum Occupancy.} 
    The points with an elevation between $0.5$ and $1.5$m are classified as medium vegetation, and the remaining points as high vegetation.
    The points are then projected onto the raster corresponding to their predicted stratum, and the raster's pixels are classified as vegetation if it contains the projection at least one such point.
\end{itemize}

\paragraph{\bf Deep Learning Baseline}
We also train a simple PointNet network to directly predict the three strata occupancy values directly. The network then follows the same training procedure than our proposed approach. See Appendix~\ref{sec:pointnet_our} for more details about this method.   
\subsection{Results}

\begin{table}
    \begin{center}
    \caption{\textbf{Quantitative results.} We report the accuracy of the predicted aggregated plot occupancy, along with the inference speed in number of plots per second.}
    \label{tab:stats}
        \begin{tabular}{lcccccc} 
        \toprule
        \multirow{2}{*}{Method}
        &\multicolumn{3}{c}{Absolute error, \%}
        &
        Inference Time  \\\cline{2-5}
        & low & medium & high & average & plt/s \\\hline
        Handcrafted & 21.9 & 20.7 & 10.3 & 17.6& 20\\
        PointNet Baseline & 17.4 & \bf 13.5 & 7.7 & 12.8 & \bf 400\\
        Ours &\bf 15.5 & 13.6 & \bf 7.5 & \bf 12.2 & 125\\\bottomrule
        \end{tabular}
    \end{center}
\end{table}

In \tabref{tab:stats}, we present the quantitative performance of the different methods evaluated. Our method outperforms the simple deep learning baseline at the cost of added computation time. Our method provides further improvements:
(i) the occupancy maps can be easily visualized for each stratum in raster form;
(ii) the point heights are explicitly modeled, so each point contributes only to the coverage of their stratum.

\begin{figure*}
\captionsetup[subfigure]{justification=centering}
    \begin{subfigure}{.5\textwidth}
    \centering
    \includegraphics[width=0.8\textwidth]{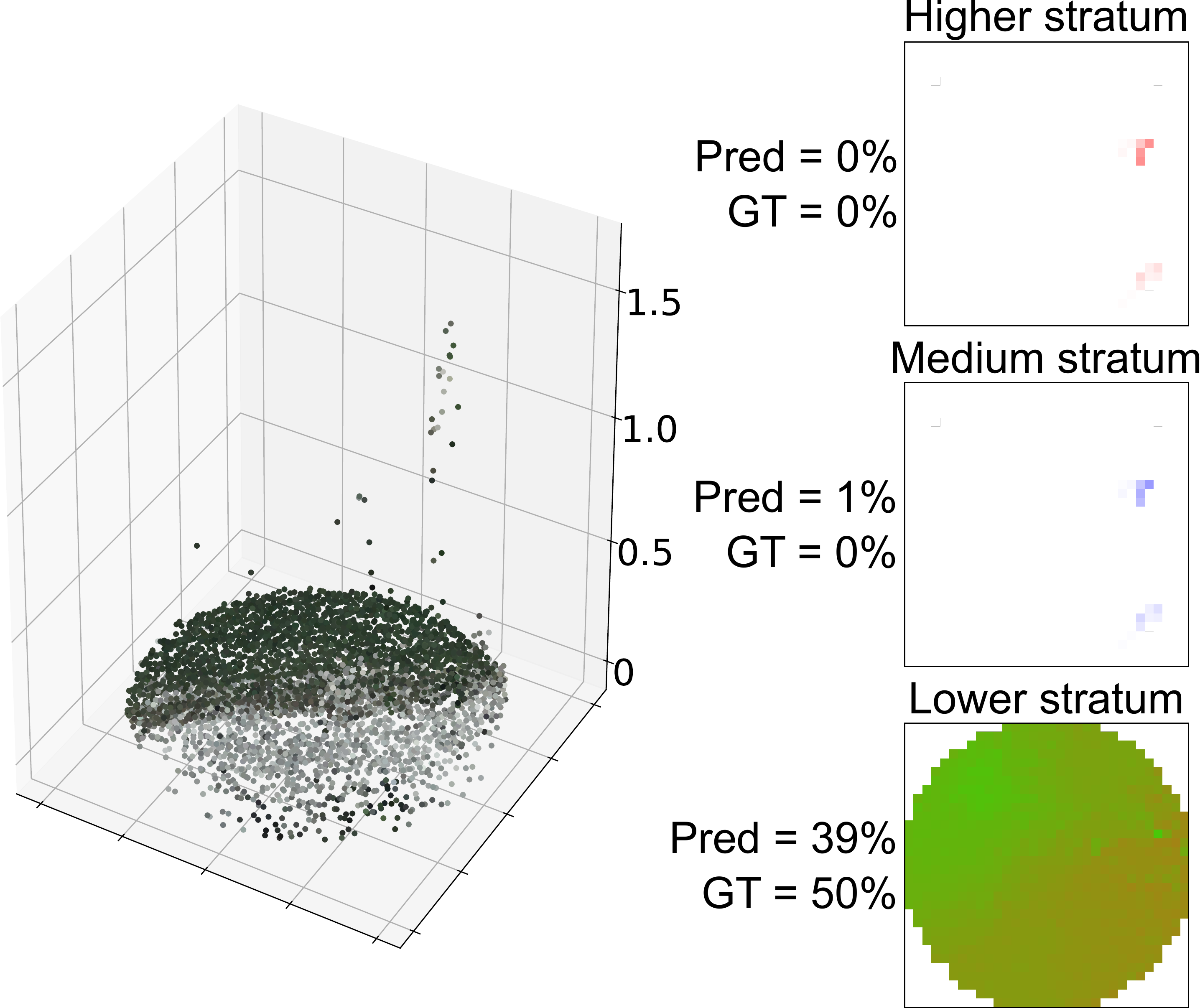}
    \caption{Simple plot without vegetation.}
    \label{fig:example_results_a}
    \end{subfigure}
    \hspace{1cm}
    \begin{subfigure}{.5\textwidth}
    \includegraphics[width=0.8\textwidth]{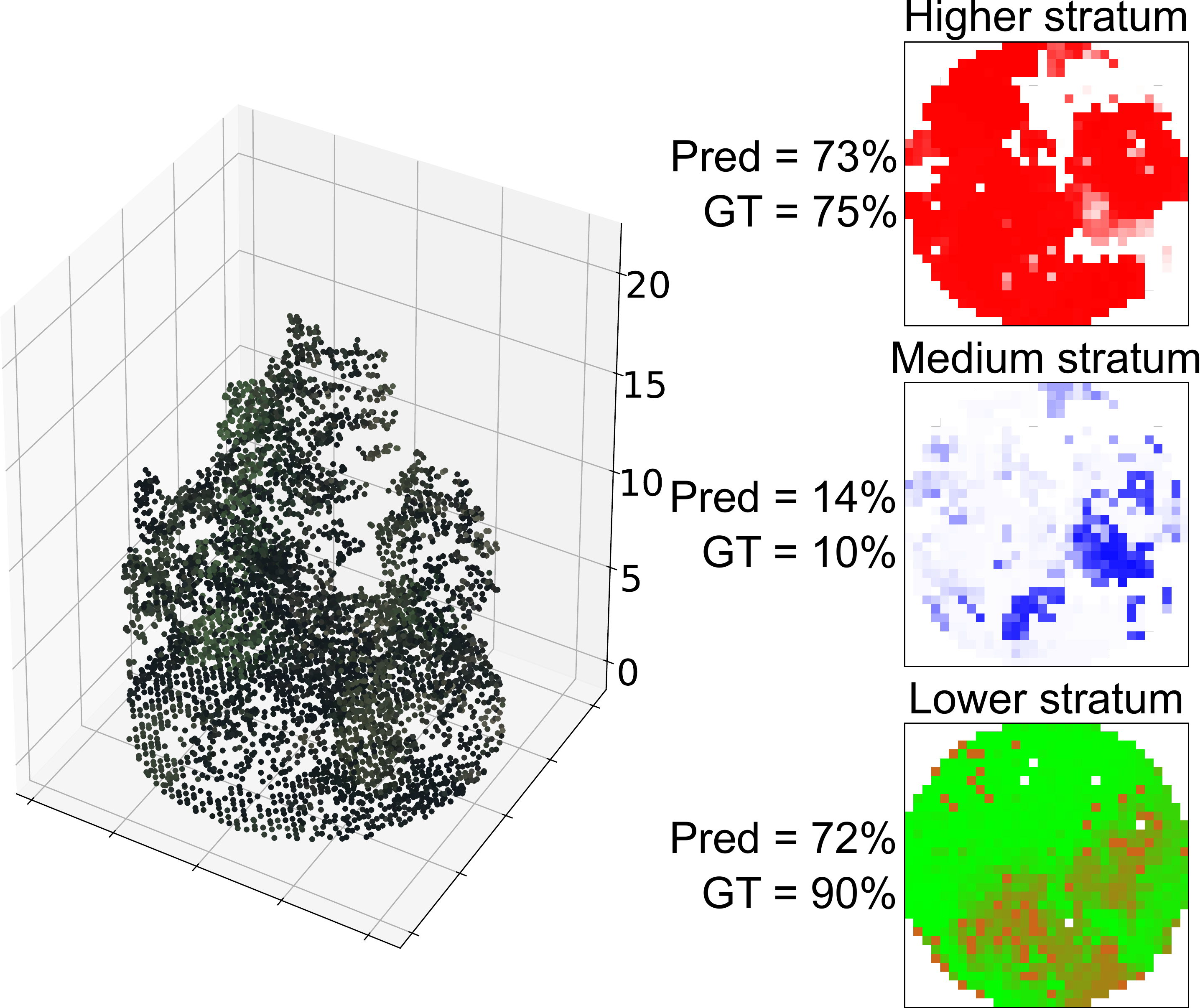}
    \caption{Complex, multi-strate plot.}
    \label{fig:example_results_b}
    \end{subfigure}
    \caption{\textbf{Qualitative Results.} Our method predicts aggregated stratum occupancy along with the corresponding rasterized occupancy maps. Here, the pixels are colored according to the value of the predicted occupancy: shades of green, blue, and red indicate pixels with high-predicted vegetation coverage for the lower, medium, and higher strata respectively, while brown corresponds to bare soil.}
    \label{fig:qualitative}
\end{figure*}

In Figure~\ref{fig:qualitative}, we present qualitative results. Despite the absence of ground truth to evaluate the quality of the predicted occupancy maps, one can see the visual correspondence between the point clouds and their corresponding strata coverage.

\begin{figure}
    \centering
    \includegraphics[width=0.7\textwidth]{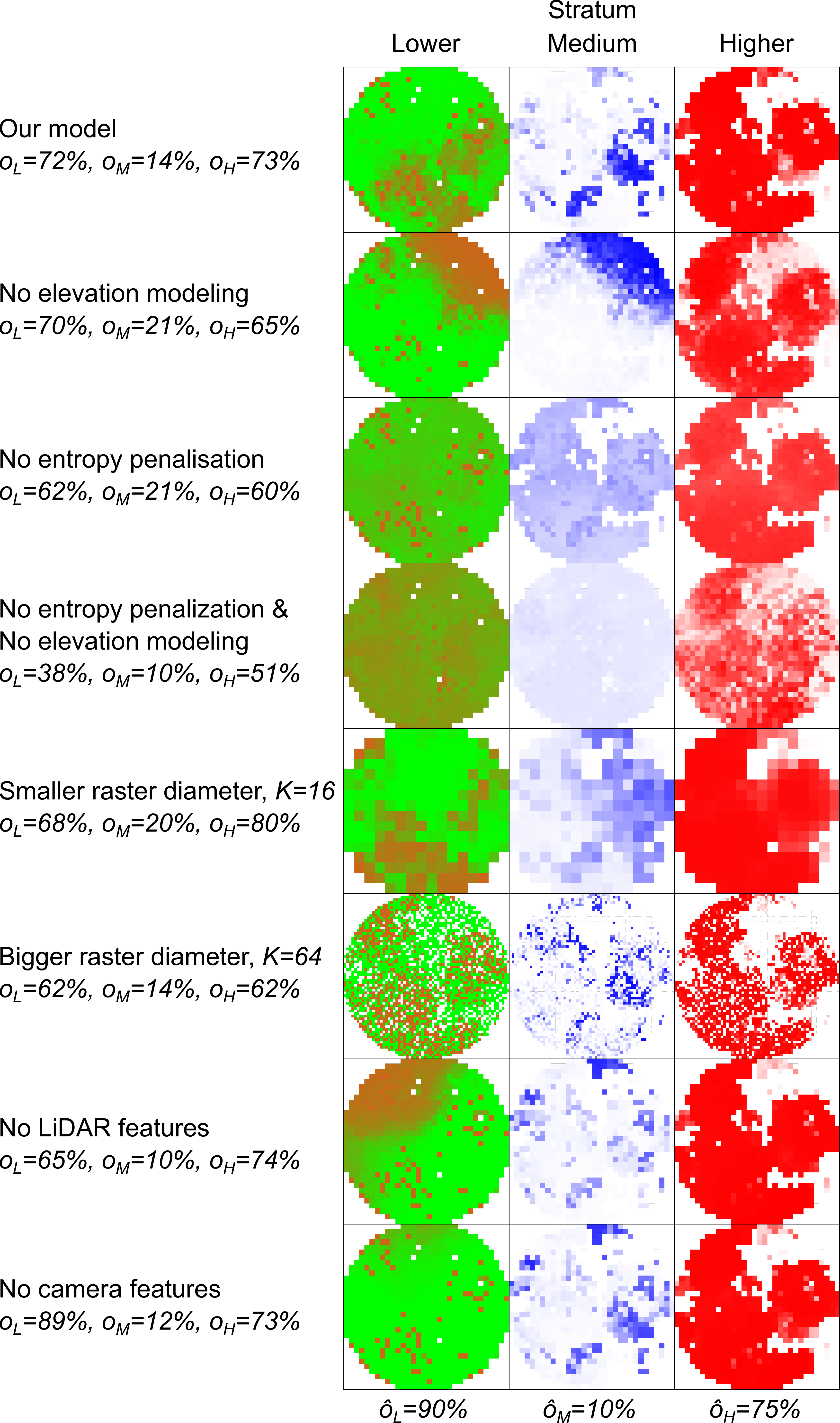}
    \caption{\textbf{Qualitative Ablation Study.} Occupancy maps produced by variations of our model, for the original point cloud presented in Figure~\ref{fig:example_results_b}. The accuracy of lower and medium stratum occupancy maps directly depends on the model configuration. However, the higher stratum occupancy map is almost identical for most models, as the higher vegetation is easiest to distinguish.}
    \label{fig:ablation_study}
\end{figure}

\label{sec:ablation_study}
In \tabref{tab:ablation}, we evaluate the quantitative impact of some of our main design choices and report quantitative results in \tabref{tab:ablation} and qualitative results in \figref{fig:ablation_study}.

\begin{table}
\begin{center}
\caption{\textbf{Quantitative Ablation Study.} Impact of some design choice on the mean absolute error.}
\label{tab:ablation}
\begin{tabular}{lcccc}
\toprule
    \multirow{2}{*}{Method}
    &\multicolumn{4}{c}{Absolute error, \%}
    \\\cline{2-5}
    & low & medium & high & average
    \\\hline
    \textbf{Our method} &\textbf{15.5} & \textbf{13.6} & \textbf{7.5} & \textbf{12.2}\\
    No elevation modeling &16.6 &13.8 &7.4 &12.6 \\
    No entropy penalisation &15.6 &13.9 &7.3 &12.3 \\
    No el. modeling\& ent. pen. &15.8 &13.4 &6.5 & \bf 11.9 \\
    Coarser raster &16.7 &14.4 &8.3 &13.1 \\
    Finer raster &15.9 &18.5 &7.1 &14.0 \\
    No LiDAR features &16.8 &13.4 &7.6 &12.5 \\
    No radiometric features &16.5 &13.4 &7.7 &12.5 \\
\bottomrule
\end{tabular}
\end{center}
\end{table}

We highlight the importance of elevation modeling and entropy penalization by presenting the performance without $\mathcal{L}_\text{elevation}$ (No elevation modeling), without $\mathcal{L}_\text{entropy}$  (No entropy penalization), and with neither (No el. modeling \& ent. pen.). We observe that both penalizations have little effect on the performance.
This is an expected results, as $\mathcal{L}_\text{data}$ directly represents the error rate. However, as illustrated in \figref{fig:ablation_study}, these regularizations play a crucial role in obtaining realistic occupancy maps. Without elevation modeling, the localization of high occupancy pixels are decorelated with the actual position of vegetation. In the absence of entropy minimization, the maps become fuzzy and lack sharp features.

We study the influence of the raster size $K$ with $K=16$ (Coarser raster) and $K=64$ (Finer raster) instead of the chosen $K=32$. We can see that the results are quantitatively worse with these choices of $K$. Lower values for $K$ produce less informative occupancy maps, while higher values lead  to many empty pixels as the number of points is not sufficient to propagate the occupancy. 

\begin{figure}
    \centering
    \includegraphics[width=0.8\textwidth]{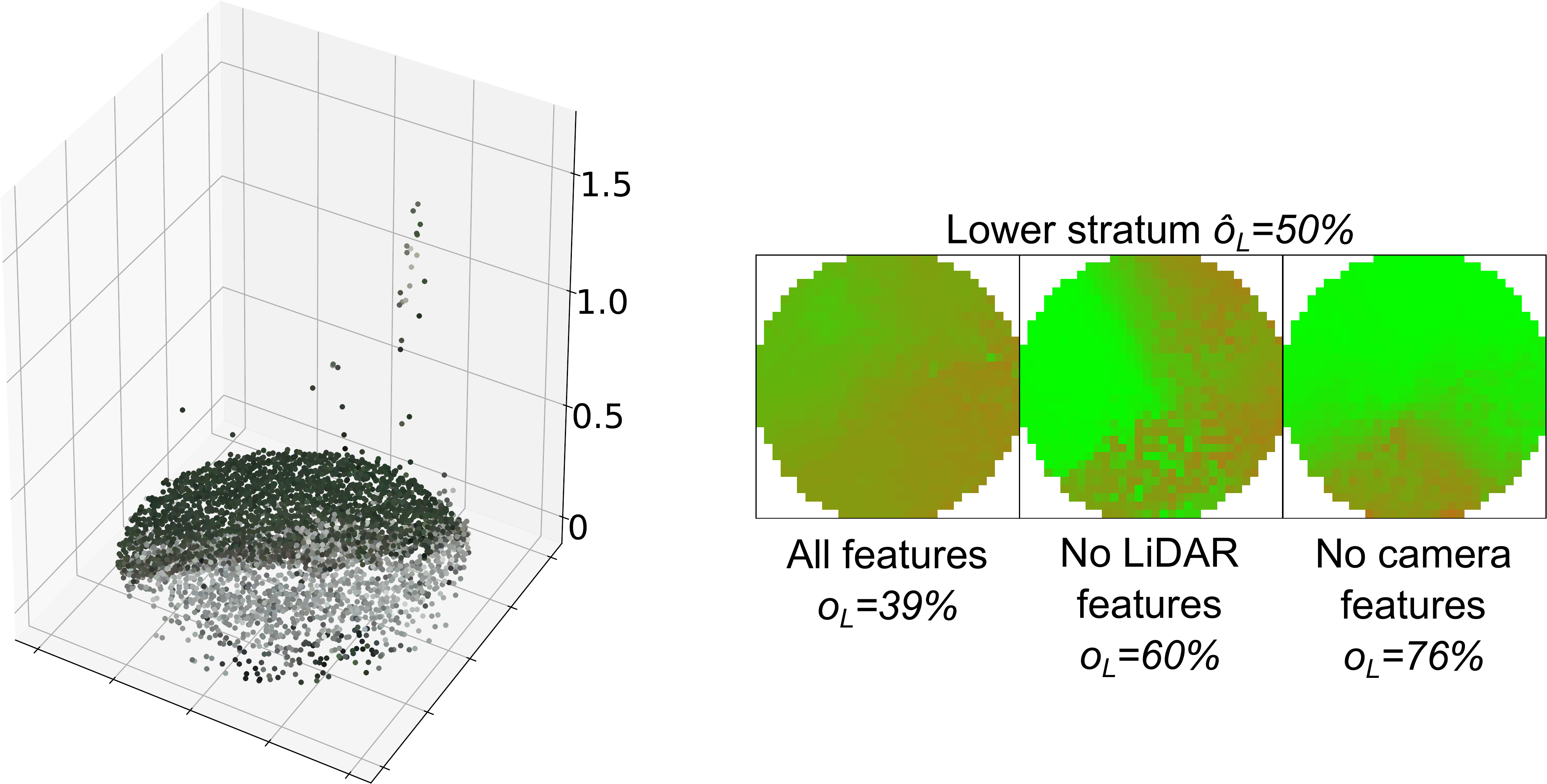}
    \caption{\textbf{Illustrative Example.} Lower stratum occupancy maps produced by our model using different sets of features for the plot in \ref{fig:example_results_a}. We observe that our original model produced a more accurate occupancy map, confirming the importance of using all features.}
    \label{fig:ablation_study_2}
\end{figure}

Finally, we evaluate the impact of non-geometric LiDAR features by removing only removing the number of returns and LiDAR reflectance (No LiDAR features), and by removing RGB and NIR (No radiometric features). Radiometric features prove useful to distinguish between low vegetation from bare soil, or leaves from tree branches. However, the benefits of these features vary from one plot to the other due to the geometry of the acquisition: the radiometry is acquired as an optical image and superposed on the 3D point cloud. Occlusions can prevent the precise colorization of points in lower stratum, see Figure~\ref{fig:ablation_study_2}.
LiDAR features, on the other hand, are not affected by the plot geometry. However, they contain less discriminative information.
In both cases, using only one set of features slightly decreases the overall model accuracy and often produces visually incorrect lower and sometimes medium stratum occupancy maps. 

\paragraph{\bf Limitations} 
While we can assess the precision of our approach in terms of aggregated predictions, we are not able to evaluate quantitatively the pixel-wise or point-wise prediction. Indeed, we do not have access to these costly annotations. Nevertheless, some plots can be easily visually validated. For example, Figure~\ref{fig:ablation_study_2} provides an example of a plot without medium or higher vegetation, and a lower stratum evenly divided into lower vegetation and bare soil. Even though none of the aggregated predictions $o_\LL$ are exact, we observe that our model produces an occupancy map with the best visual fidelity to the plot. Hence, our model is able to combine all available features for better occupancy map prediction.

The global loss function used to train the model is composed of three different losses, which implies some manual parameter tuning. However, as the weights of the networks are learned from data, our method does not rely on handcrafted parameters, which typically require setting numerous parameters by hand. As our data only originates from one region, we are not able to evaluate the robustness of our approach. However, since our method is very generic in its formulation, it should be able to be trained on data from another area and not require manual tuning.

Finally, the ECM algorithm for elevation modelling requires a manual initialization step. However, the parameters of the Gamma distributions are intuitive as they relate to the moment of the distribution (mean height, deviation) and can be approximated by a knowledgeable operator. Alternatively, this can be hand-tuned using the likelihood as a guide, and only needs to be approximately tuned before running the optimization.

\section*{Conclusion}
\label{sec:conclusion}
In this paper, we presented a 3D deep learning method for predicting occupancy across three vegetation strata: lower, medium, and higher.
Using only three aggregated values per example plot, our model is able to perform a point-wise
classification and produce vegetation occupancy rasters with a high precision and at a small computational cost.
Moreover, our model was 30\% more accurate and 6 times faster than a handcrafted approach.
Our code is released in open access, along with the first forestry dataset with occupancy annotations.

\section*{Acknowledgments}
\label{sec:acknowledgments}
This study has been co-funded by CNES (TOSCA FRISBEE Project, convention $n^o 200769/00$) and CONFETTI Project (Nouvelle Aquitaine Region project, France).


\FloatBarrier
\bibliography{biblio}

\clearpage
\renewcommand{\thesection}{A.\arabic{section}}
\section*{Appendix}
\setcounter{section}{0}


\section{PointNet Model Baseline}
\label{sec:pointnet}
Our algorithm is based on PointNet model for semantic segmentation \cite{pointnet}. Given a point cloud $X \in \bfR^{N \times D_0}$, where $D_0$ is the number of point features, the model predict the scores $P=\{p_0, ..., p_N\} \in \bfR^{N \times C} $ that a point in $X$ belongs to each of $C$ classes.

The model process the data in the following manner:
\begin{itemize}
    \item To homogenise the input data, we first subsample each point cloud in $M$ points (\eqref{eq:sampling}). 
    \item The first MultiLayer Perceptron (MLP) $\MLP_1 : D_0 \mapsto D_1$ is applied to each point $i=1 \cdots M$ in parallel and maps raw point features to a learned point descriptors $f_i^1$ of size $D_1$. This MLP is composed of a sequence of 1D convolutional layers \cite{Goodfellow-et-al-2016}, batch normalization \cite{pmlr-v37-ioffe15}, and Rectified Linear Units (ReLU) \cite{relu} (\eqref{eq:mlp1}).
    \item $\MLP_2 : D_1 \mapsto D_2$ operates in the similar way (\eqref{eq:mlp2}).
    \item The maxpooling operation is applied over $M$ points to extract a global shape descriptor (GSD) $G$ of size $1 \times D_2$ for each point cloud (\eqref{eq:maxpool}).
    \item $G$ is concatenated with the output of $\MLP_1$ block, then each point is processed by $\MLP_3 (D_1+D_2) \mapsto C$ to extract the classes predictions as in previous MLP blocks. Note, that ReLU activation and BatchNorm are not applied to the last layer (\eqref{eq:mlp3}).
    \item Finally, we upsample the obtained predictions to $N$ points by using the nearest neighbour algorithm (\eqref{eq:upsampling}). 
\end{itemize}

\begin{figure}[ht!]
    \centering
    \includegraphics[scale=0.7]{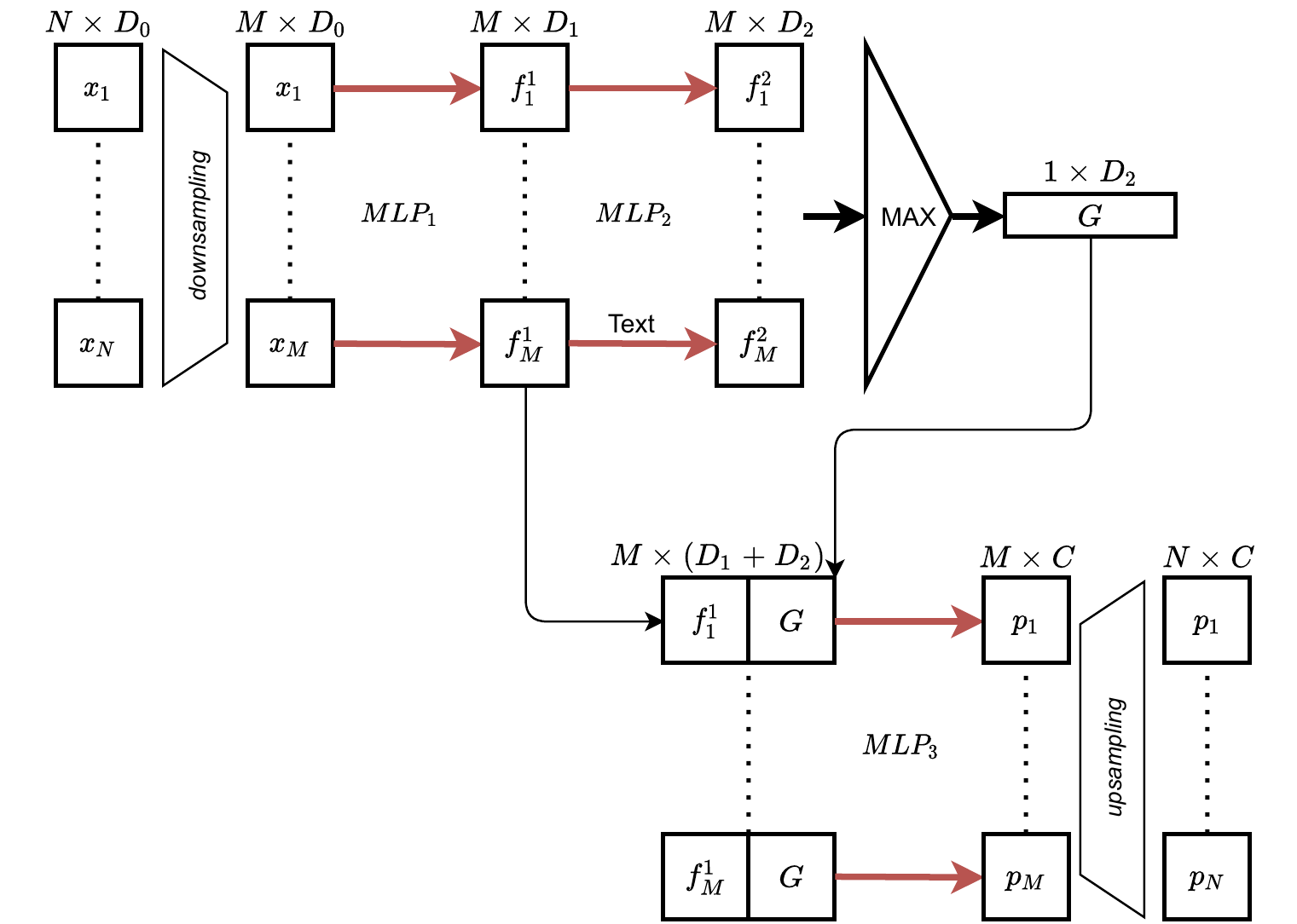}
    \caption{{\bf PointNet model for Semantic Segmentation \cite{pointnet}}. The preprocessing step is consists in data subsampling from $N$ to $M$ points with $D_0=9$ features. $\MLP_1$ and $\MLP_2$ consecutively extract features $F_1$ and $F_2$ from 3D point cloud data, afterwards the maxpooling operator produces a global shape descriptor $G$. It is then concatenated with the first set of extracted features $F_1$, which are processed by $\MLP_3$ to extract the final predictions $P$ for $C$ classes. In the postprocessing step, we upsample the point cloud to its original size of $N$ points.}
    \label{fig:their_pointnet}
\end{figure}

The PointNet model for semantic segmentation is presented on Figure~\ref{fig:their_pointnet} and can be summarized by the following equations:

\begin{align}
    \label{eq:sampling}
    M &= \text{sample}(M, N) ,\\
    \label{eq:mlp1}
    {f_m^1} &= \MLP_1(x_m),\; \forall m \in M ,\\
    \label{eq:mlp2}
    {f_m^2} &= \MLP_2(f_m^1),\; \forall m \in M,\\
    \label{eq:maxpool} 
    G &= \text{MAX}(F_2), \; F_2=\{f_1^2, ..., f_M^2\},\\
    \label{eq:mlp3}
    {p_m} &= \MLP_3(\left[f_m^1 \mid\mid G\right]),\; \forall m \in M,\\
    \label{eq:upsampling}
    N &= \text{upsample}(N, M) ,
\end{align}
where $\left[. \mid\mid .\right]$ is the concatenation operator.

\section{PointNet Baseline}
\label{sec:pointnet_our}
We compare our method to a simple deep learning baseline in which the PointNet network is used to directly predict the different layer occupancies from raw observations. We consider a point cloud $X \in \bfR^{N \times 9}$ with $N$ points each characterized by the $9$ radiometric and geometric point features described in \secref{sec:data}.

\paragraph{Architecture}
As in the original PointNet model we first subsample the point cloud in $M$ points that are then processed by a $\MLP_1: \bfR^9 \mapsto D$ to extract $D$ point features $f_i$, $i\in[1, ..., N]$. In the following step, we extract a GSD $G$. Finally, the second MLP $\MLP_2 : \bfR^D \mapsto 3$ maps the GSD to three values defining the predicted stratum occupancy $o_\LL$, $o_\ML$ and $o_\HL$. Contrary to $\MLP_1$, $\MLP_2$ is composed of linear layers paired with ReLU activation for all the layers, except the last one that uses the Sigmoid activation~\cite{10.1007/3-540-59497-3_175}.


\begin{figure}[h]
    \centering
    \includegraphics[scale=0.8]{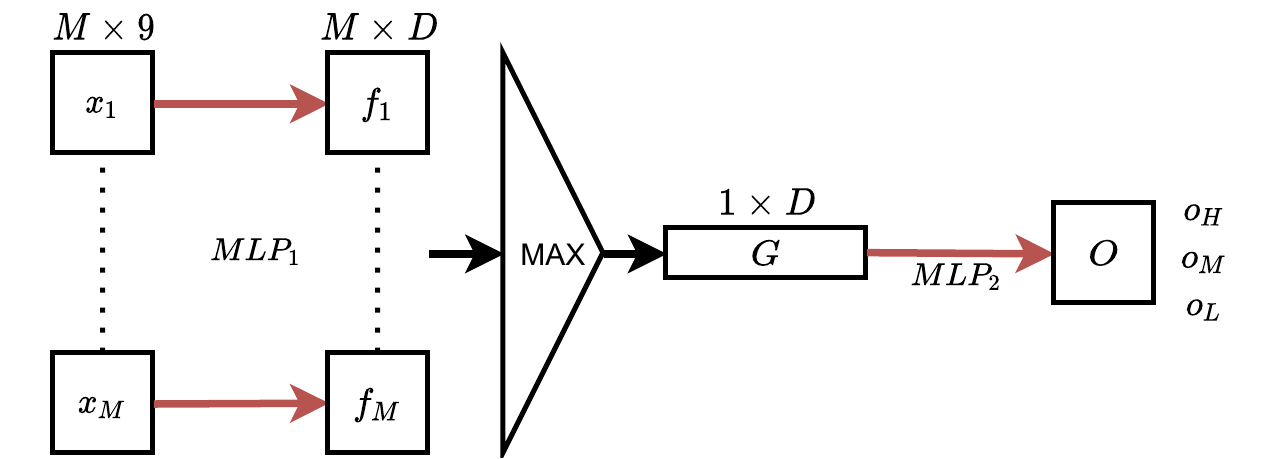}
    \caption{Schematic view of our adapted PointNet model. The subsampled point cloud is processed by $\MLP_1$ to extract a set of point features $F$, then the maxpooling operator produces a global shape descriptor $G$. Finally, $\MLP_2$ is applied to $G$ to extract the stratum occupancy values $o_\LL, o_\ML, o_\HL$.}
    \label{fig:our_pointnet}
\end{figure}

\paragraph{Loss Function} We use the same $\ell_1$ norm loss function as for our model (\eqref{eq:loss_l1}).

\paragraph{Implementations Details} The PointNet baseline is trained with a batch size of $20$, the ADAM \cite{kingma2017adam} optimizer with a learning rate of 0.001 divided by 10 every 50 epochs, and default parameters.
We add a dropout layer \cite{JMLR:v15:srivastava14a} with probability $0.4$ before the last layer to prevent overfitting of the model.
The output sizes of MLP blocks of the model are respectively: $[32,32,64,128]$ and $[64, 32]$. The number of subsampled points for each plot is $M=2048$.

\section{Bayesian Elevation Modeling}

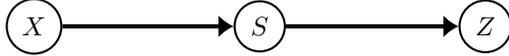
\begin{figure}
    \centering
    \begin{tikzpicture}
    \node[thick, draw=black, circle] (nx) at (0,0) {$X$};
    \node[thick, draw=black, circle] (ns) at (3,0) {$S$};
    \node[thick, draw=black, circle] (nz) at (6,0) {$Z$};
    
    \draw[-{Latex[length=2mm, width=3mm]}, ultra thick] (nx) -- (ns);
    \draw[-{Latex[length=2mm, width=3mm]}, ultra thick] (ns) -- (nz);
    \end{tikzpicture}
    \caption{Graphical model representation of the proposed probabilistic model linking the observation $X$, the ground/non-ground status of points $S$, and the elevation $Z$. By using a generative model for the elevation of a point conditionally to its stratum, we can quantify the compatibility between a stratum prediction and its corresponding elevation.}
    \label{fig:gm}
\end{figure}

We propose a probabilistic model linking the following random variables: the raw observation $X \in \mathbf{R}^{N \times 9}$, the ground/non-ground nature of each point $S \in \{\G, \NONG\}^N$,  and $Z \in \mathbf{R}^N_+$ the points' elevations.
The conditional dependencies between these variables are represented in \figref{fig:gm}. We propose to consider $S$ the ground / non-ground variable as a discrete latent variable connecting the observation and the elevation.\footnote{Note that since the elevation is already comprised in the observation, adding the arbitrary notion of stratum as an intermediate variable does not provide any rich insight in terms of modeling. It allows us however to link in a natural manner the stratum prediction and the elevation distribution.}

We can write the likelihood of the observed elevation $z \in \mathbf{R}^N_+$ conditionally to the observations according to this model:
\begin{align}
\ell(z)
= \prod_{n=1}^N 
    P(z_n \mid X)
= \prod_{n=1}^N
      P(z_n, S=\G \mid X)
    + P(z_n, S=\NONG \mid X)~.
    \end{align}
After applying Bayes theorem, we obtain:
\begin{align}
\ell(z)
&= 
\prod_{n=1}^N
        P(S=\G \mid X) P(z_n \mid X,  S=\G)
      + P(S=\NONG \mid X) P(z_n \mid X,  S=\NONG)~,
    \end{align}
and using the conditional independence between $Z$ and $X$ with respect to $S$, we have that:
\begin{align}
\ell(z)
&= \prod_{n=1}^N
        P(S=\G \mid X) P(z_n \mid S=\G)
      + P(S=\NONG \mid X) P(z_n \mid S=\NONG)~.
\end{align}
The conditional elevation distributions $P(z_n \mid S=\G)$ and $P(z_n \mid S=\G)$ are parameterized by $\Gamma_\G$ and $\Gamma_\NONG$ respectively. The posterior probabilities of belonging to the ground or nonground stratum are given by the network's prediction: $P(S=\G \mid X) = Y_\LL + Y_\BS$ and $P(S=\NONG \mid X) = Y_\ML + Y_\HL$. Finally, we obtain that:
\begin{align}
\ell((z)
&= \prod_{n=1}^N
        (Y_\LL + Y_\BS) \Gamma_\G(z_n)
      + (Y_\ML + Y_\HL) \Gamma_\NONG(z_n)~.
\end{align}
We define the loss $\ell_\text{elevation}$ as the negative log-likelihood of the observed elevations conditionally to the observations:
\begin{align}
\mathcal{L}_\text{elevation}
&=
    -\log\left(\ell((z)\right) \\
    &=-\sum_{n=1}^N \log\left(
     (Y_\LL + Y_\BS) \Gamma_\G(z_n)
      + (Y_\ML + Y_\HL) \Gamma_\NONG(z_n) \right)~.
\end{align}

\end{document}